\documentclass[letterpaper]{article} %
\usepackage{aaai24}  %
\usepackage{times}  %
\usepackage{helvet}  %
\usepackage{courier}  %
\usepackage[hyphens]{url}  %
\usepackage{graphicx} %
\usepackage{natbib}  %
\usepackage{caption} %
\usepackage{algorithm}
\usepackage{algorithmic}

\usepackage{subcaption}
\usepackage{booktabs}
\usepackage{multirow}

\usepackage{newfloat}
\usepackage{listings}
\DeclareCaptionStyle{ruled}{labelfont=normalfont,labelsep=colon,strut=off} %
\floatstyle{ruled}
\newfloat{listing}{tb}{lst}{}
\floatname{listing}{Listing}
\title{MUDD: A New Re-Identification Dataset with Efficient Annotation for Off-Road Racers in Extreme Conditions}
\author {
    Jacob Tyo\textsuperscript{\rm 1,2}\equalcontrib,
    Motolani Olarinre\textsuperscript{\rm 1}\equalcontrib,
    Youngseog Chung\textsuperscript{\rm 1},
    Zachary C. Lipton\textsuperscript{\rm 1}
}
\affiliations {
    \textsuperscript{\rm 1}Carnegie Mellon University\\
    \textsuperscript{\rm 2}DEVCOM Army Research Laboratory\\
    \{ jtyo, young, tolani, zlipton \}@cs.cmu.edu
}

\begin{document}

\maketitle

\begin{abstract}
Re-identifying individuals in unconstrained environments 
remains an open challenge in computer vision. 
We introduce the Muddy Racer re-IDentification Dataset (MUDD), 
the first large-scale benchmark 
for matching identities of motorcycle racers 
during off-road competitions. 
MUDD exhibits heavy mud occlusion, 
motion blurring, complex poses, 
and extreme lighting conditions 
previously unseen in existing re-id datasets. 
We present an annotation methodology 
incorporating auxiliary information 
that reduced labeling time by over 65\%. 
We establish benchmark performance 
using state-of-the-art re-id models 
including OSNet and ResNet-50. 
Without fine-tuning, 
the best models achieve only 33\% Rank-1 accuracy. 
Fine-tuning on MUDD boosts results to 79\% Rank-1, 
but significant room for improvement remains.
We analyze the impact of real-world factors 
including mud, pose, lighting, and more. 
Our work exposes open problems 
in re-identifying individuals under extreme conditions. 
We hope MUDD serves as a diverse
and challenging benchmark to spur progress 
in robust re-id, 
especially for computer vision applications 
in emerging sports analytics.
All code and data can be found at \url{https://github.com/JacobTyo/MUDD}.
\end{abstract}

\section{Introduction}

Re-identifying individuals across disjoint camera views 
is a fundamental task in computer vision. 
Despite progress, 
most research assumes controlled capture environments 
and consistent appearances~\citep{gou2018systematic}. 
When restricted to such controlled environments,
existing solutions do a good job 
of handling challenges due to occlusion, 
pose variation, and lighting changes.
although further progress is needed~\cite{ye2021deep}. 
However, as we show, %
outside of such controlled environments,
current techniques struggle.

\begin{figure}[t]
    \centering
    \includegraphics[width=0.25\textwidth]{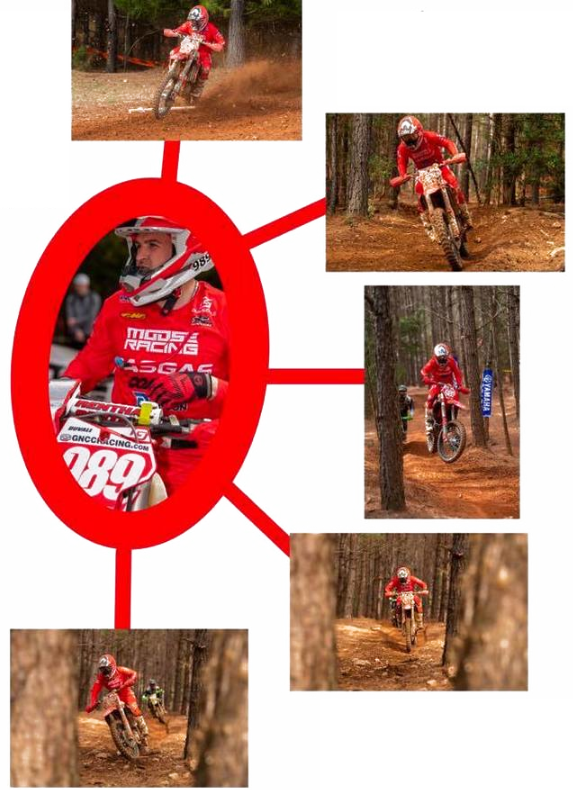}
    \caption{Motorcycle Racer Re-Identification}
    \label{fig:cover-photo}
\end{figure}

In particular, we focus on the task of
identifying motorcycle racers 
during off-road competitions (Figure \ref{fig:cover-photo})
through mud, dirt, trees, and crowds. 
Here, appearances can change drastically 
lap-to-lap as mud accumulates or subsequently flies off. 
Numbered jerseys that could 
otherwise be used to easily re-id racers
often become obscured by mud, 
are out of sight of the camera, or get torn. 
Glare, blurring, and extreme lighting also occur
as a single racing event can go from bright fields to 
deep dark forests. 
To the best of our knowledge, 
there exists no public datasets
to supports research into robust 
re-id under such conditions.

To spur progress in addressing these challenges, 
we introduce the Muddy Racer Re-Identification Dataset (MUDD). 
MUDD contains $3,906$ images 
of $150$ identities captured over
$10$ off-road events 
by $16$ professional motorsports photographers. 
The imagery exhibits heavy mud occlusions, 
complex poses, distant perspectives, motion blurring, and more. 
We also present an efficient annotation methodology 
incorporating detected racer numbers as auxiliary information 
to generate high-quality identity clusters for manual verification. 
This improved labeling time by over 65\% 
compared to more simplistic labeling methods.

We establish benchmark performance using state-of-the-art 
re-id models based on a Omni-Scale CNN Neural Network~\citep{zhou2019omni} 
and ResNet-50~\citep{he2016deep}. 
Without fine-tuning, 
the best models reach only $22\%$ Rank-1 accuracy. 
But when fine-tuning is incorporated, 
the best models reach $79\%$ Rank-1 accuracy. 
Interestingly, 
pretraining
can be performed with ImageNet data~\citep{deng2009imagenet} 
to achieve nearly identical performance 
as 
pretraining
on re-identification (re-id) specific datasets. 
Despite this increase in performance, 
a considerable gap remains between machine and human performance. 

Our analysis reveals open problems in handling mud occlusion, 
appearance changes, poses, resolution, and similar outfits. 
These factors induce intra-class variation 
and inter-class similarity 
that current models fail to robustly distinguish.
In summary, we introduce a diverse, 
challenging dataset exposing the limitations 
of existing re-id techniques. 
MUDD provides imagery to drive progress 
in re-identification amidst uncontrolled real-world conditions.

Our contributions are:
\begin{itemize}
    \item The MUDD dataset containing diverse imagery 
    to evaluate re-id of off-road racers. 
    To our knowledge, 
    this represents the first large-scale dataset 
    of this emerging application domain.
    \item A method to improve 
    annotation effectiveness 
    by incorporating auxiliary information during labeling.
    \item Initial benchmarking of state-of-the-art models, 
    which reveal limitations on this dataset 
    and substantial room for further improvement.
    \item Analysis of failure cases which provide insights 
    to guide future research on robust re-identification 
    for sports analytics and computer vision broadly.
\end{itemize}

\begin{figure}[t]
    \includegraphics[width=0.5\textwidth]{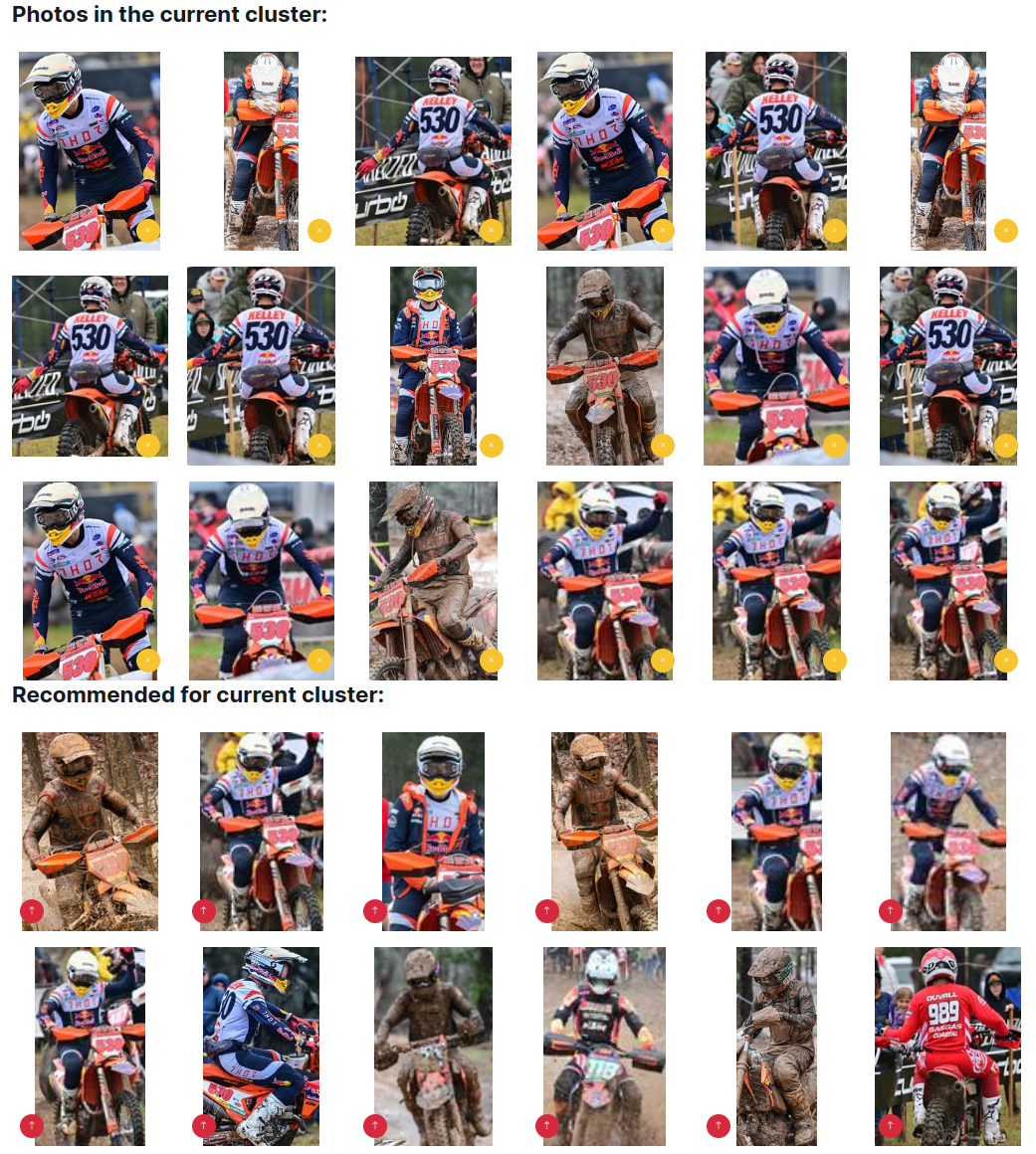}
    \caption{Leveraging detected jersey numbers as auxiliary information enables generating higher quality identity clustering proposals for manual verification. This proposed cluster contains both clean and muddy images of the same rider, whereas proposing clusters with off-the-shelf re-id models fail.}
    \label{fig:label_example1}
\end{figure}

\begin{figure}[t]
    \centering
    \includegraphics[width=0.5\textwidth]{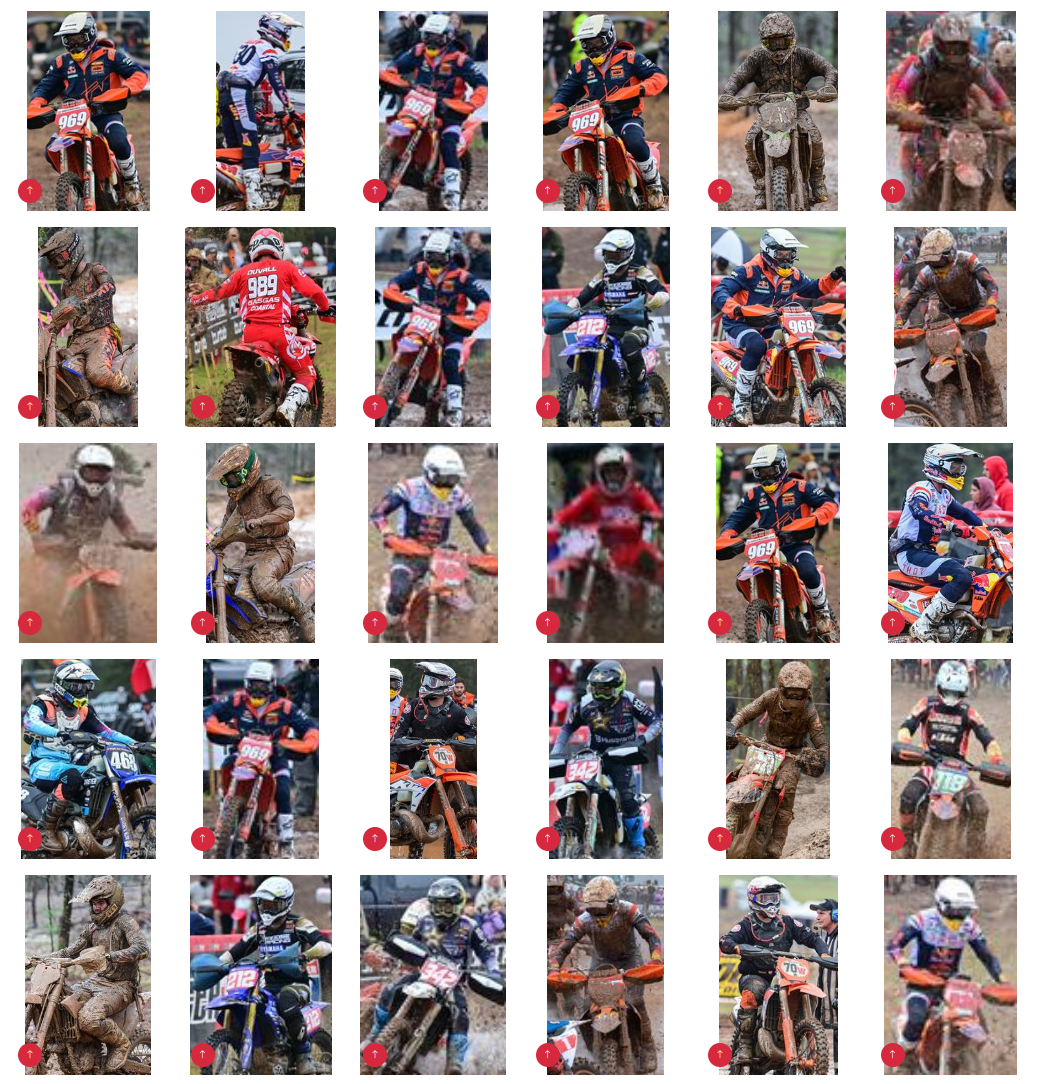}
    \caption{Additional proposed results for the same identity cluster as Figure~\ref{fig:label_example1}. Our methodology provides high-quality recommendations to simplify manual verification and labeling.}
    \label{fig:label_example2}
\end{figure}

\section{Efficient Labeling via Auxiliary Information}
\label{sec:auxinfo}

One key challenge in constructing re-id datasets 
is to efficiently group images 
of the same identity during labeling. 
Exhaustively labeling identities from scratch can become
intractable for a large dataset of images 
with an unknown number of identities. 
To assist in this labeling process, 
images can be clustered 
into groups using pretrained models,
and then manually verified by annotators. 
However, in constructing MUDD, 
we found that annotators still spent over 30 
minutes on each identity, 
requiring a more efficient process.

Off-the-shelf re-id models focus 
on extracting features invariant 
to nuisance factors like pose, 
lighting, and blurring
while discriminating between identities. 
However, these features are based on their pretraining dataset
and they cannot explicitly 
leverage domain-specific cues---especially
if the domain-specific cues are not available 
in the pretraining dataset, such as racer numbers. 
The re-id model treats 
the image holistically 
without localizing and 
recognizing semantic concepts like digits.
Therefore, when the models are applied 
on different image domains, 
any useful domain-specific cues are not used.

\begin{table*}[t]
\centering
\caption{MUDD re-id benchmark results comparing off-the-shelf, from scratch, and fine-tuning training strategies. Fine-tuning provides major accuracy gains indicating the importance of transfer learning.}
\label{tab:results}
\begin{tabular}{@{}cccccc@{}}
\toprule
Training                    & Backbone  & R1 & R5 & R10 & mAP \\ \midrule
\multirow{2}{*}{Best off-the-shelf}            & OSNet     & 0.3252 & 0.5219 & 0.6327 & 0.3853 \\
                                        & ResNet-50 & 0.3164 & 0.5099 & 0.6306 & 0.3634 \\ \midrule 
\multirow{2}{*}{Trained From Scratch}        & OSNet &  0.2146 (0.03108)  &  0.4846 (0.04466)  &  0.6755 (0.03606)   &   0.2491 (0.0163)  \\
                            & ResNet-50    &  0.1591 (0.01154)  &  0.4155 (0.01564)  &  0.6194 (0.2826)   &  0.1923 (0.01853)   \\ \midrule
\multirow{2}{*}{Pretrained on Imagenet}      & OSNet &  0.7844 (0.01284)  &  0.9416 (0.005594)  &   0.9771 (0.004829)  &  0.8215 (0.01258)   \\
                            & ResNet-50    &  0.762 (0.00817)  &  0.9442 (0.004079)  &  0.9787 (0.002729)   &  0.8073 (0.006272)   \\
\multirow{2}{*}{Pretrained on MSMT17}     & OSNet &  0.7924 (0.009929)  &  0.9445 (0.001521)  &  0.9779 (0.002051)   & \textbf{0.8287 (0.005843)}    \\
                            & ResNet-50    &  0.7596 (0.02279)  & 0.9407 (0.0118)   &  0.9767 (0.006813)   &   0.8028 (0.02378)  \\ 
\multirow{2}{*}{Pretrained on DukeMTMC}     & OSNet &  0.7887 (0.01515)  &  0.9388 (0.003319)  &  97.57 (0.004367)   & 0.826 (0.0117)    \\
                            & ResNet-50    &  0.7858 (0.01726)  &  \textbf{0.9562 (0.007937)}  &  \textbf{0.9847 (0.002225)}   &   0.8277 (0.01079)  \\
\multirow{2}{*}{Pretrained on Market-1501}     & OSNet &  \textbf{0.7931 (0.01738)}  &  0.9442 (0.006051)  &  0.9778 (0.00483)   &  0.827 (0.01546)   \\
                            & ResNet-50    &  0.7812 (0.02546)  &  0.9475 (0.01395)  &  0.9807 (0.008821)   &   0.8233 (0.02077)  \\ \bottomrule
\end{tabular}
\end{table*}

In light of this challenge, 
we leverage the fact that each identity (i.e. racer) 
in this dataset is assigned a visible number 
and we propose directly utilizing this auxiliary information 
during the clustering and re-id process via 
a pretrained text detection model~\citep{lyu2018mask}.
This domain knowledge provides 
strong localization cues 
to group images with the same numbers. 
The re-id model alone struggles 
to consistently spot 
and match the small digit regions 
amidst mud, motion, and variations.

Explicitly guiding search and clustering with the auxiliary numbers, 
even if noisy, complements the holistic re-id model. 
Our breadth-first attribute search 
leverages the domain knowledge 
to effectively explore the data 
and retrieve number matches. 
This creates high-quality initial clusters 
that seed the depth-first re-id search.

In essence, we get the best of both worlds:
domain-driven localization from the auxiliary cues, 
combined with holistic identity discrimination from the re-id model. 
The re-id model alone lacks 
the explicit semantic guidance, 
resulting in poorer search and clustering. 
Our hybrid approach better utilizes 
both domain knowledge and learned representations.

Specifically,
to generate ground truth labels for specific racers, 
we first extract all numbers using a
pretrained text detection model~\citep{lyu2018mask}, 
and also create a re-id embedding using a pretrained OSNet model.
Then we iterate over the following process:
\begin{enumerate}
    \item Pick a number that was detected more than 10 times and retrieve all images containing it.
    \item For each result from Step 1, 
    take the top $k$ nearest neighbors based on the re-id embedding.
    \item Combine the results for each search by rank, 
    and present %
    to annotators for manual refinement and verification.
\end{enumerate}
This updated process reduced the average time 
to verify an identity cluster 
from over 30 minutes to under 10.

Figure~\ref{fig:label_example1}
shows a proposed cluster from our labeling system.
The top section contains all photos where the number 
530 was detected. 
The bottom section shows the most similar images
according to the pertained OSNet re-id model. 
Critically, leveraging the auxiliary number information provides 
an initial cluster with clean and muddy images 
of the same racer that can be used as a seed image 
for a search by the re-id model. 
Figure \ref{fig:label_example2} shows additional results deeper in the ranking.

\section{The MUDD Dataset}

MUDD\footnote{MUDD is available at \url{https://github.com/JacobTyo/MUDD}} contains $3906$ images 
capturing $150$ identities
across 10 different off-road events from 
the Grand National Cross Country (GNCC) racing series. 
The events span various track conditions, 
weather, 
times of day, 
and racing formats. 
Images were captured by 16 professional motorsports photographers 
using a diverse range of high-end cameras.

We gathered a large library of off-road competition photos 
from the off-road photography platform \url{PerformancePhoto.co}. 
We used YOLOX~\citep{ge2021yolox} 
to detect the bounding boxes for people. 
An embedding was extracted for each cropped bounding box using 
the general-use re-id model OmniScaleNet~\citep{zhou2019omni}. 
We leverage a scene text spotter~\citep{lyu2018mask} 
to extract visible racer numbers 
as auxiliary information to aid our labeling process, 
as detailed in Section 2.
Importantly, 
the accuracy of the auxiliary models on MUDD data is low. 
Our scene text spotter has less than 50\% end-to-end accuracy. 
However, as described in Section 2, 
even low-accuracy auxiliary information 
can still drastically improve annotation 
efficiency by enabling effective search and clustering.

\begin{figure*}[t]
    \centering
    \includegraphics[width=\textwidth]{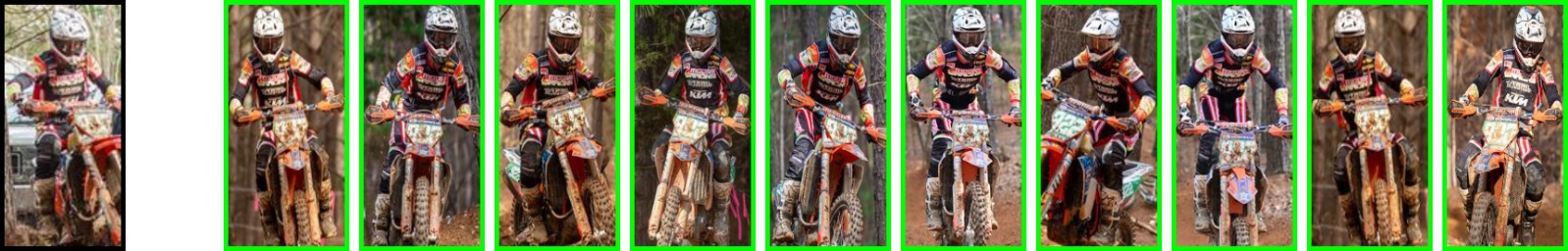}
    \caption{Example of successful re-id by the fine-tuned model under moderate mud occlusion. The 10 top retrievals correctly identify the query rider despite mud, pose, and other variations. Green boundaries signify correct matches and red incorrect.}
    \label{fig:lightmud}
\end{figure*}

\begin{figure*}[t]
    \centering
    \includegraphics[width=\textwidth]{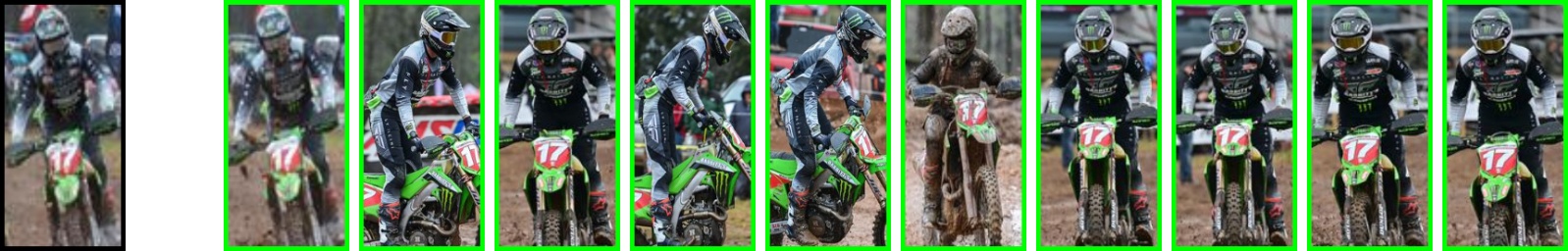}
    \caption{Example of the model correctly matching a clean image of a rider to a muddy image of the same rider when the pose is similar between the query and gallery image. Green boundaries signify correct matches and red incorrect.}
    \label{fig:clean-to-muddy}
\end{figure*}

We manually labeled all identities, 
accelerated by our proposed method. 
Some individuals occur in multiple events, 
either with very similar outfits or entirely different ones. 
To simplify training and evaluation, 
we provide an event ID and treat the same individual 
across events as different identities.

MUDD contains several major challenges:%
\begin{itemize}
    \item Heavy mud occlusion - 
    Racers accumulate significant mud spatters 
    and caking. %
    This represents a unique occlusion pattern not present 
    in existing re-id datasets.
    \item Complex poses---Racers exhibit varied poses including leaning, 
    jumping, crashes, and more unseen during regular walking.
    \item Distance and resolution---Images captured from a distance with small, 
    low-resolution racer crops.
    \item Dynamic lighting---Outdoor conditions cause glare, 
    shadows, and exposure variations.
    \item Clothing---Jerseys and numbers that could ease 
    re-id are often obscured by mud, gear, and positioning.
    \item Motion blur---Racers maneuver at high speeds causing 
    motion-blurring effects, 
    especially combined with panning cameras.
\end{itemize}

We divided MUDD into train (80\%) and test (20\%) sets. 
There is no identity overlap between the sets. 
We further divided the train set into 
a train and validation split also with a 90/10 ratio. 
The validation set was used for model selection, 
hyperparameter tuning, 
and ablation studies. 
All metrics reported on the held-out test set.

The dataset includes identities under a 
variety of motorcycle and riding gear. 
It captures both professional 
and amateur events across multiple 
states during the first 7 months of 2023. 
The diversity of identities, 
environments, 
perspectives, 
and conditions 
exceeds existing re-id datasets.

\begin{figure*}[t]
    \centering
    \includegraphics[width=\textwidth]{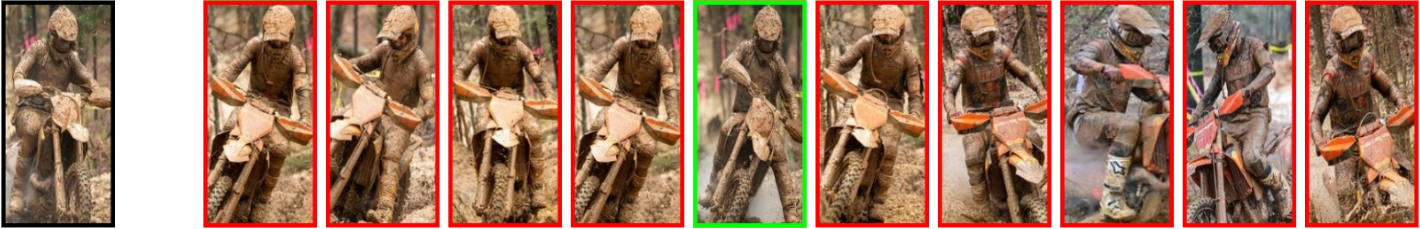}
    \caption{Failure case with heavy mud occlusion on the query image. Only 1 out of the top 10 results is a correct match, despite over 20 images of the same rider appearing in the gallery set, most of which are clean. Green boundaries signify correct matches and red incorrect.
    }
    \label{fig:very-muddy-query}
\end{figure*}

\begin{figure*}[t]
    \centering
    \includegraphics[width=\textwidth]{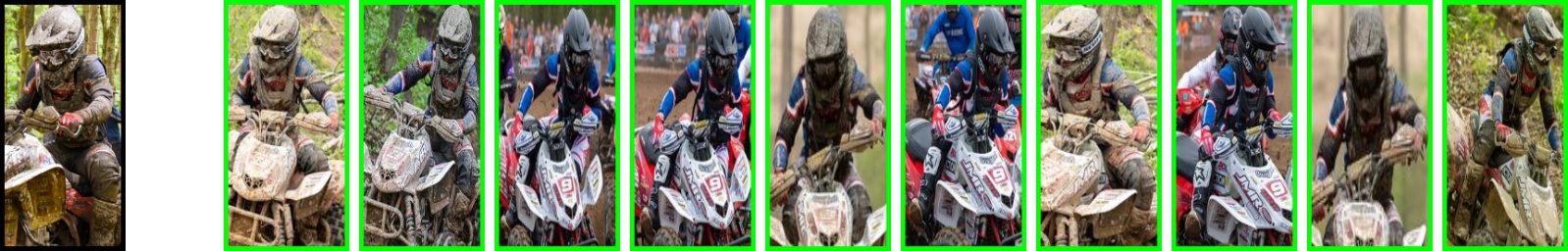}
    \caption{Example of successful re-id by the fine-tuned model under light mud occlusion. All top 10 ranked results correctly match the query rider despite mud, blurring, lighting, pose, and complex backgrounds. Green boundaries signify correct matches and red incorrect.}
    \label{fig:light-mud-success}
\end{figure*}

\begin{figure*}[t]
    \centering
    \includegraphics[width=\textwidth]{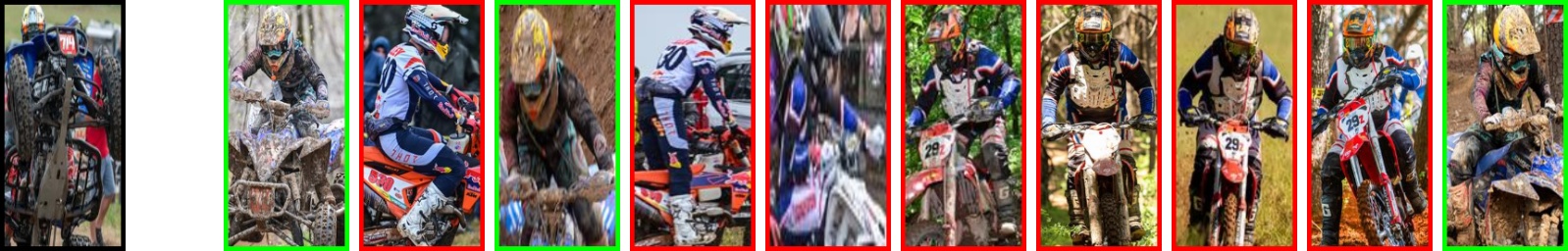}
    \caption{Example of a failure case due to extreme pose variation in the query image. The rider is captured doing a wheelie, leading to incorrect matches despite no mud occlusion. Green boundaries signify correct matches and red incorrect.}
    \label{fig:wheelie-pose}
\end{figure*}

\begin{figure*}[t]
    \centering
    \includegraphics[width=\textwidth]{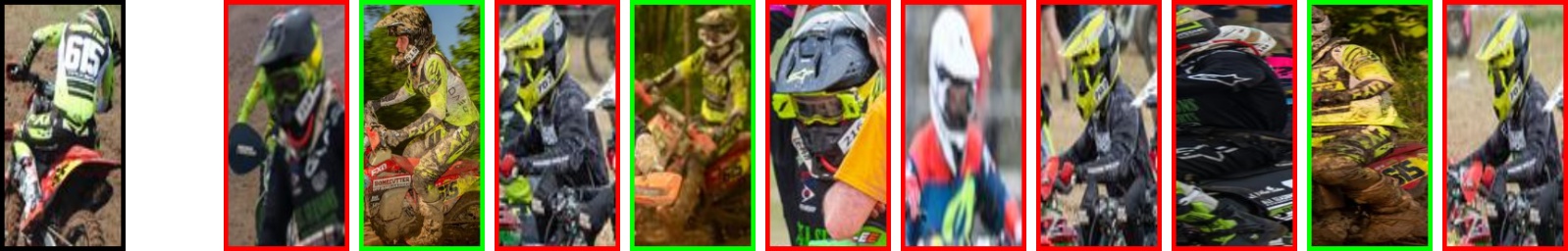}
    \caption{Failure case due to pose variation between the query and gallery images. The backward-facing query rider is not matched to forward-facing images of the same identity. Green boundaries signify correct matches and red incorrect.}
    \label{fig:poses-are-hard}
\end{figure*}

\begin{figure*}[t]
    \centering
    \includegraphics[width=\textwidth]{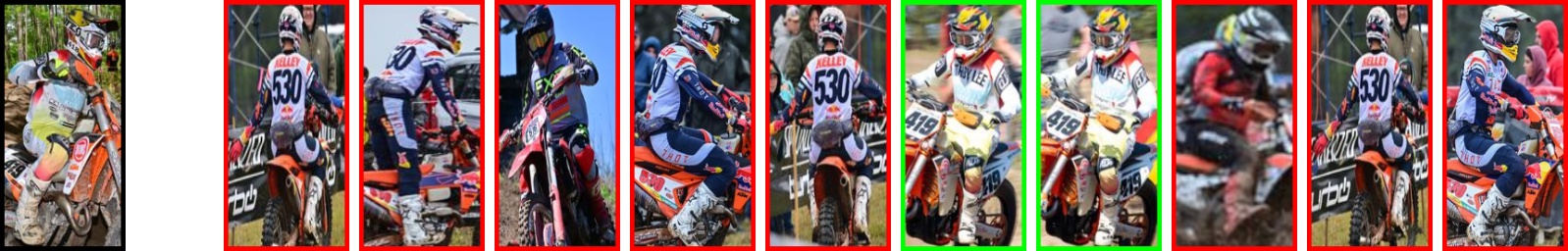}
    \caption{Example failure case due to two different riders having very similar jerseys and gear, leading to confusion between their identities. Green boundaries signify correct matches and red incorrect.}
    \label{fig:similar_outfits}
\end{figure*}

\begin{figure*}[t]
    \centering
    \includegraphics[width=\textwidth]{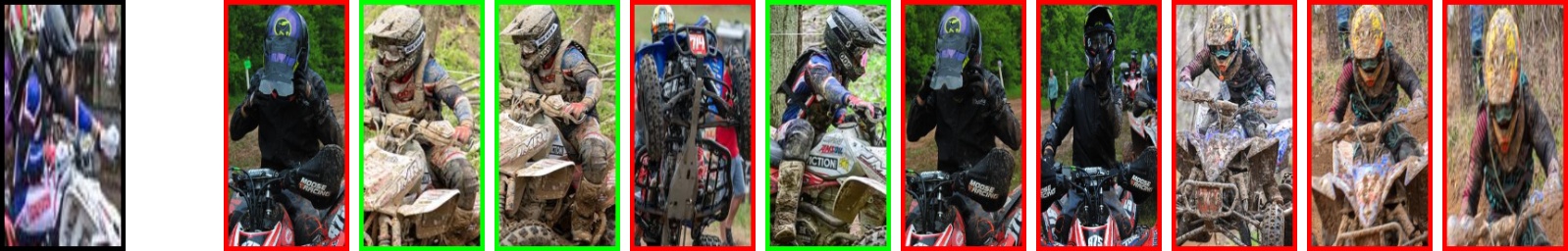}
    \caption{Failure case due to low resolution of the query image preventing distinguishing details from being visible. The small, distant crop of the rider cannot be matched accurately. Green boundaries signify correct matches and red incorrect.}
    \label{fig:lowres}
\end{figure*}

\section{Experiments}

We evaluated the performance of models on MUDD in three settings:

\begin{itemize}
\item \textbf{Off-the-shelf}: Pre-trained state-of-the-art re-id models applied directly to MUDD. 
\item \textbf{Random Initialization}: Models trained from random initialization only on MUDD. 
\item \textbf{Transfer}: Person re-identification pre-trained models 
fine-tuned on MUDD. 
\end{itemize}

We selected strong open-source implementations of 
CNN-based architectures, hereafter
referred to as OSNet~\citep{zhou2019omni}
and ResNet50~\citep{he2016deep}.

For training, 
we used triplet loss, 
and data augmentation of random flips, 
color jitter, and random crop. 
Models were optimized using Adam.
We tuned hyperparameters like learning rate, 
batch size, 
and data augmentation techniques 
based on the validation set.
All models were trained for 100 epochs,
using a cosine learning rate schedule with a 
maximum learning rate 
of $0.0003$.
The final performance is reported on the test set
at the best checkpoint, 
and all models were trained on a single NVIDIA 2080Ti GPU. 
The mean and standard deviation are reported over three
random seeds. 

\subsection{Evaluation Metrics}

We adopt the standard re-id metrics 
cumulative matching characteristic (CMC) 
rank-1, rank-5, rank-10
and mean Average Precision (mAP). 
CMC measures rank-$k$ accuracy, 
the probability of the true match appearing in the top $k$. 
The mAP metric computes mean average precision across all queries. 
Both operate directly on the re-id model output.

\section{Results}
\label{sec:results}

Table \ref{tab:results} summarizes re-id 
performance on MUDD.

\paragraph{Off-the-shelf Models} 
Applying pre-trained re-id models directly 
to MUDD leads to very poor accuracy. 
The highest Rank-1 is only 32.52\% using OSNet pre-trained on Market-1501. 
This highlights the significant domain gap between existing re-id datasets 
and MUDD's challenging conditions. 
Off-the-shelf models fail to generalize.

\paragraph{Training from Scratch} 
Begining with a random initialization, training models directly on MUDD 
struggles to learn effectively. 
OSNet is only able to achieve 21.46\% Rank-1 accuracy, 
indicating the difficulty of learning a robust representation from this training data alone. 

\paragraph{Fine-tuning}
Fine-tuning pre-trained models 
by resuming optimization on MUDD 
significantly improves accuracy. 
Fine-tuned OSNet reaches 79.31\% Rank-1, 
over 2.5x higher than off-the-shelf and 3.7x higher 
than training from scratch. 
Fine-tuning transfers invariant features 
and discrimination capabilities 
from larger source datasets, 
allowing models to adapt to MUDD 
despite the limited training data.

Interestingly, 
models pre-trained on generic ImageNet data 
perform nearly as well as those pre-trained on 
re-id specific datasets like Market-1501 after fine-tuning. 
This indicates MUDD represents a significant domain shift 
even from existing re-id datasets. 
The ImageNet features still provide a useful initialization 
for fine-tuning to this new domain.

\paragraph{Architectures} 
We experimented with two CNN-based architectures: 
OSNet, specifically designed for re-id tasks \cite{zhou2019omni}, 
and ResNet-50, 
a general-purpose CNN also commonly used for re-id \cite{he2016deep}. 
After fine-tuning on MUDD, 
OSNet achieves slightly higher Rank-1 accuracy (79.31\%) than ResNet-50 (78.12\%).

This performance gap may stem from OSNet's 
specialized representations 
tailored for scale-invariance on people. 
In contrast, 
ResNet's more general features still perform competitively, 
demonstrating the versatility of standard CNNs. 
Overall, 
both architectures can adapt to MUDD's domain when fine-tuned, 
with OSNet's inductive biases providing a small boost. 
However, 
substantial room for improvement remains compared to human performance.

\paragraph{Pretraining Datasets} 
We considered models already tailored to the 
person re-identification task.  
Starting with models pretrained on one of the 
re-id datasets of
MSMT17~\citep{wei2018person}, 
DukeMTMC~\citep{ristani2016performance}, 
or Market-1501~\citep{zheng2015scalable}, 
we fine-tune the models further on MUDD. 
The performance of these models is comparable 
across different source datasets, 
all substantially improving over off-the-shelf 
and from scratch approaches. 

In summary, 
pre-training provides significant accuracy gains 
by overcoming the limited training data through transfer learning. 
However, gaps to human-level performance remain, 
motivating techniques tailored to MUDD's extreme conditions. 
The results reinforce the dataset's unique challenges 
and domain shift from existing re-id datasets.

\section{Analysis}

Our fine-tuned models demonstrate significant improvements 
in re-identifying riders compared to off-the-shelf 
and from-scratch approaches. 
As seen in Figures~\ref{fig:lightmud} 
and \ref{fig:light-mud-success}, 
the model is able to correctly match identities 
even with mud occlusion if the rider's pose is relatively consistent. 
This indicates that fine-tuning successfully 
incorporates invariances to mud 
while still distinguishing small inter-class 
differences like gear and outfit.

However, challenges remain under more extreme conditions. 
In the rest of this section, 
we analyze several key factors 
that still cause fine-tuned model failures on MUDD:

\paragraph{Mud occlusion} 
As expected, 
heavy mud occlusion poses significant challenges. 
Mud induces high intra-class variation 
as the amount of mud covering a rider 
can vary drastically across images. 
It also causes low inter-class variation 
since mud occludes distinguishing features 
like jersey numbers and colors. 
As shown in Figure \ref{fig:very-muddy-query}, 
querying with a muddy image 
retrieves other muddy images 
rather than cleaner images of the same identity.

\paragraph{Appearance variation} 
Natural appearance changes of a rider 
over a race also confuses models. 
Riders may change gear like goggles 
or gloves multiple times. 
Crashes can rip clothing and jerseys. 
The model must learn to link different levels of mud, 
gear, and damage of a rider.

\paragraph{Pose variation} 
Complex poses like jumps, crashes, 
and wheelies are difficult to match, 
especially combined with mud and appearance variation. 
As seen in Figure \ref{fig:wheelie-pose}, 
a rider doing a wheelie is not matched 
to more standard riding poses. 
Even common pose differences like 
front versus back views 
are challenging (Figure \ref{fig:poses-are-hard}).

\paragraph{Low resolution} 
Images with small, 
distant crops of riders lack fine details 
for discrimination. 
Figure \ref{fig:lowres} 
shows a failure case where the query is low resolution.

\paragraph{Similar outfits} 
In some cases, 
different riders with very similar 
gear are confused. 
This is common as racers supported by the same 
team will typically purposefully 
coordinate their appearance. 
An example is shown in 
Figure \ref{fig:similar_outfits}.

In summary, 
heavy mud occlusion, 
appearance changes, 
pose variations, 
low resolution, 
and similar outfits remain open challenges. 
While fine-tuning offers substantial improvements, 
significant gaps compared 
to human performance motivates the need for 
new techniques tailored to these 
uncontrolled conditions.

\section{Limitations}

While MUDD enables new research 
into re-id under extreme conditions, 
our work has several limitations to note:

\paragraph{Labeling Bias} 
Our accelerated labeling methodology leveraging 
auxiliary information could introduce bias. 
By searching for images matching the same detected number, 
we preferentially sampled identities with more visible numbers. 
Not only may this over-represent riders with cleaner jerseys 
and under-represent heavy mud occlusion, 
but also many riders choose to have very little numbering. 
The labeling distribution may not fully reflect the underlying data. 
Models could overfit to the biases of our annotation process. 
Collecting additional labeled data with different 
sampling approaches could help quantify and reduce this bias.

\paragraph{Dataset Size} 
MUDD contains 3,906 images across 150 identities. 
While large for this emerging domain, 
this remains small compared 
to widely used re-id datasets. 
The limited data makes learning robust models difficult. 
Additional identities and examples 
would likely improve accuracy,
but scaling dataset size is costly 
in this domain.

\paragraph{Capture Bias} 
All MUDD data was captured during 
the first half of 2023 across 10 events 
in the GNCC racing series. 
This induces bias in the environments, 
rider identities, and more. 
Performance may not transfer 
to other off-road competitions 
like motocross, supercross, and flat track events. 
Broader capture diversity could improve model robustness. 

\paragraph{Camera IDs} 
MUDD lacks camera ID labels denoting 
which images came from the same capture device. 
Camera ID is a useful cue for re-id, 
enabling models to account 
for consistent environmental 
factors and biases per device. 
However, our dataset combines imagery 
from 16 different independent photographers 
at unknown shooting locations.

\section{Related Work}

Person re-identification (re-id) 
aims to match people across 
non-overlapping camera views and time 
horizons~\citep{ye2021deep, zheng2016person, zheng2015scalable, farenzena2010person}. 
Early re-id methods relied on handcrafted features 
like color histograms, textures, 
and local descriptors~\citep{farenzena2010person}. 
With the rise of deep learning, 
Convolutional Neural Network (CNN)~\cite{zhou2019omni} 
and Transformer~\citep{he2021transreid} 
based approaches now dominate re-id research, 
spurred by datasets like Market-1501~\citep{zheng2015scalable}, 
DukeMTMC-ReID~\citep{ristani2016performance}, 
and MSMT17~\citep{wei2018person}.

A few datasets address environmental factors. 
For example, 
\citet{xiao2016end} introduce a dataset with a low-resolution 
challenge set. 
Occlusions have also 
been well studied, 
spearheaded by datasets with high levels of occlusion~\citep{schwartz2009learning, wang2011re, wang2016person, figueira2015hda+, xiao2016end}.
However, 
these occlusions are unrelated
to the heavy mud occlusion in our dataset. 
The addition of mud drastically 
complicates re-identification.
Furthermore, 
no prior datasets exhibit such a complex combination of lighting, 
diversity, motion, and diverse cameras as 
our off-road racing dataset. 

Prior work has focused on the re-identification of motorcycles and bicycles~\cite{figueiredo2021more, li2022rider, yuan2018bike}, 
however these are restricted to street vehicles in urban settings.
A highly related domain is identifying athletes in sports imagery.
\citet{PENATESANCHEZ2020355} release a dataset of ultra-runners
competing in a 128km race over the course of a day and a night. 
While this is more similar to the off-road setting in our dataset, 
they only have 416 different identities between 5 locations at a single event. 
Furthermore, 
there is near zero mud in the dataset.
Along similar lines, 
but in even more controlled and limited settings, 
are the SoccarNet-ReID~\citep{giancola2022soccernet} 
and DeepSportRadar-ReID~\citep{van2022deepsportradar} datasets, 
which contain images from broadcast video 
of soccer and basketball games respectively. 

These datasets have driven research to develop 
methods to deal with the occlusions common in them. 
Approaches such as 
invariant representations~\citep{chen2019learning}, 
metric learning~\citep{yi2014deep}, 
semantic attributes~\citep{shi2015transferring},
part-based~\citep{cheng2016person} 
and pose-aware models~\cite{cho2016improving}, 
and adversarial learning~\citep{huang2018adversarially}
have been proposed to alleviate occlusion problems. 
Other methods have been developed to 
handle misalignment,
utilizes temporal cues in video~\citep{li2019global},
use domain adaptation techniques~\citep{deng2018image},
or unsupervised methods~\citep{fan2018unsupervised}
to reduce label dependencies. 
Unlike our dataset, 
these all operate in controlled conditions.
Existing models thus fail on our data. 

In summary, 
re-id research has focused 
on controlled conditions 
and modest variation. 
Our dataset introduces real-world challenges 
absent in existing datasets.
Our experiments expose clear gaps 
between current methods and this %
application. 
MUDD provides diverse imagery to spur 
new techniques for robust re-id under uncontrolled conditions.

\section{Conclusion}

In this work, 
we introduce MUDD, 
the first large-scale dataset to benchmark 
re-identification of motorcycle racers under extreme conditions. 
MUDD captures challenging factors including heavy mud occlusion, 
complex poses, variable lighting, 
and distant perspectives. 
We propose an accelerated annotation methodology 
incorporating detected racer numbers %
to enable efficient high-quality labeling.

Through initial benchmarking experiments, 
we demonstrate significant gaps between current re-id techniques 
and the real-world conditions represented in MUDD. 
Off-the-shelf models fail to generalize to this new domain. 
Training CNN models like OSNet and ResNet from scratch 
struggles due to the limited training identities, but
fine-tuning pre-trained models on MUDD significantly improves accuracy.
Interestingly, 
models pre-trained on generic ImageNet data 
prove as effective as re-id-specific pre-training.

However, 
substantial gaps compared to human performance remain even after fine-tuning. 
Our analysis reveals open challenges 
including handling heavy mud occlusion, complex poses, low resolution, 
and similar outfits. 
These factors induce intra-class variation 
and inter-class similarity that current models fail 
to robustly distinguish.

In summary, 
MUDD exposes clear limitations 
of existing re-id techniques under uncontrolled conditions. 
Our work motivates new solutions tailored 
to the unique challenges of identifying 
motorcycle racers amidst mud and more. 
Broader applications such as sports analytics 
stand to benefit from progress in re-id robustness. 
MUDD provides diverse, 
real-world imagery
to drive future research towards re-identification in the wild.

\bibliography{references}

\end{document}